\pdfoutput=1
\documentclass{article}
\usepackage{spconf,amsmath,graphicx}

\usepackage{enumitem}
\setlist{nosep, leftmargin=14pt}

\usepackage{mwe} 
\usepackage[backref]{hyperref} 
\usepackage{url}
\usepackage{amsfonts}
\usepackage{bbm}
\usepackage{graphicx}
\usepackage{booktabs}
\usepackage{float}
\usepackage{multirow}
\graphicspath{{./figs/}}

\title{domain adaptive synapse detection with weak point annotations}
\name{Qi Chen$^{1}$, Wei Huang$^{1}$, Yueyi Zhang$^{1,2}$, Zhiwei Xiong$^{1,2,\dag}$
\thanks{ 
\textsuperscript{\dag} Corresponding author: zwxiong@ustc.edu.cn}}
\address{$^1$University of Science and Technology of China\\
$^2$Institute of Artificial Intelligence, Hefei Comprehensive National Science Center}

\begin{document}
%
\maketitle
\begin{abstract}
The development of learning-based methods has greatly improved the detection of synapses from electron microscopy (EM) images. However, training a model for each dataset is time-consuming and requires extensive annotations. Additionally, it is difficult to apply a learned model to data from different brain regions due to variations in data distributions. In this paper, we present AdaSyn, a two-stage segmentation-based framework for domain adaptive synapse detection with weak point annotations. In the first stage, we address the detection problem by utilizing a segmentation-based pipeline to obtain synaptic instance masks. In the second stage, we improve model generalizability on target data by regenerating square masks to get high-quality pseudo labels. Benefiting from our high-accuracy detection results, we introduce the distance nearest principle to match paired pre-synapses and post-synapses. In the WASPSYN challenge at ISBI 2023, our method ranks the 1st place. 
\end{abstract}
\begin{keywords}
Synapse detection, electron microscopy, domain adaptation, pseudo labeling
\end{keywords}
\section{Introduction}
\label{sec:intro}

Synapses are messengers in the process of transmitting information flow between neurons. Understanding synaptic connectivity is crucial for comprehending brain function and dysfunction. To locate synapses, imaging methods with a large field of view and nanometer resolution are needed. Electron Microscopy~\cite{motta2019dense,chen2022mask,huang2022semi,liu2022biological,liu2023soma} has been developed to meet these requirements, resulting in the production of terabyte and petabyte-scale image volumes. Developing highly accurate synapse detection algorithms is vital to this field.

Machine learning techniques can accurately detect synapses in large-scale images. Early approaches to machine learning focus on segmenting the synaptic cleft region using hand-crafted features~\cite{kreshuk2011automated,becker2013learning,huang2014identifying,jagadeesh2014synapse,kreshuk2014automated,plaza2014annotating}. William~\cite{roncal2014vesicle} develops a framework called Syconn, which utilizes deep convolutional neural networks and random forest classifiers to infer a richly annotated synaptic connectivity matrix by automatically identifying various components. To determine the direction of information transmission among neurons, researchers proposed the task of predicting synapse polarity. Initially, some works employ random forest classifiers for classifying pre-synapses and post-synapses ~\cite{kreshuk2015talking,staffler2017synem}, but later deep learning methods become more prevalent in this area\cite{huang2018fully,buhmann2018synaptic,heinrich2018synaptic,parag2018detecting,parag2018detecting}. Nicholas~\cite{turner2020synaptic} reframe the problem of identifying synaptic partners as directly generating the mask of the synaptic partners from a given cleft. Lin~\cite{lin2020two} introduces active learning with a query suggestion method for 3D synapse detection. The existing methods can achieve good performance in a small brain region. However, there is a wide variety of synapse textures in the brain, making it difficult to maintain consistent accuracy throughout all regions. 

In this paper, we concentrate on formulating robust algorithms for domain-adaptive synapse detection to facilitate model generalization from the source to the target brain regions. We propose a two-stage segmentation-based framework (AdaSyn) that utilizes weak point annotations to achieve its objectives. In the first stage, we first obtain the ground truth of synapse masks by expansion operation and train a 3D segmentation network (SegNet) to predict synaptic regions. SegNet outputs two channels: one for pre-synaptic regions and another for post-synaptic regions. These masks are then processed using connected component labeling to separate individual synapses. We determine the location of each synapse by calculating the center point coordinates of the corresponding masks. To match the pair of pre-synapses and post-synapses, we assign the nearest pre-synapse ID to each post-synapse using the nearest neighbor principle. To improve the generalization of our network across different brain regions, we adopt a pre-trained model on source data to generate square masks as pseudo labels for the target data. Finally, we fine-tune our SegNet using both source and target data to enhance its ability to generalize. Our method ranks 1st place on the leaderboard of the WASPSYN challenge~\cite{wu2023out}.

\begin{figure*}[t]
    \centering
    \includegraphics[width=\linewidth]{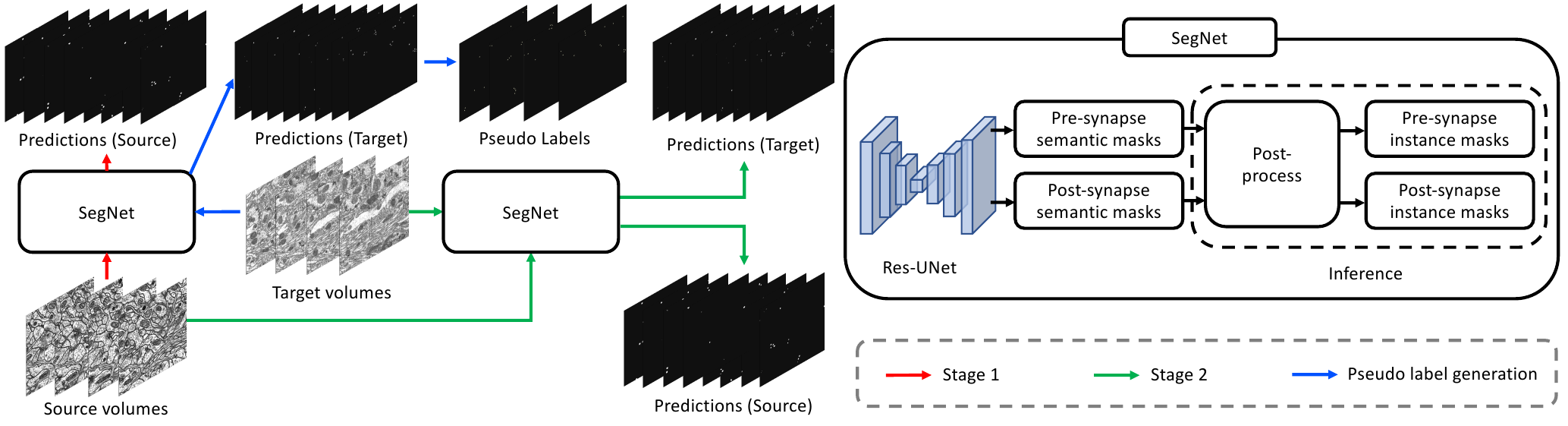} 
    \vspace{-6mm}
    \caption{The overall framework of AdaSyn. In the first stage, we train an initial segmentation network called SegNet using the source volumes as training data. In the second stage, we use the pre-trained SegNet to generate pseudo labels for the target volumes. Finally, we fine-tune the SegNet by incorporating both source labels and target pseudo labels as supervision.}
    \label{fig:framework}
\end{figure*}

\section{METHOD}
\label{sec:method}
\subsection{Problem Formulation}
We focus on the problem of unsupervised domain adaptation in synapse detection. In the source domain, we have access to source volumes
$X_{\mathcal{S}}=\left\{x_s \in \mathbb{R}^{H \times W \times D}\right\}_{s \in \mathcal{S}}$ and the corresponding pre-synapse point labels $C_{\mathcal{S}}=\left\{(c_{sx}, c_{sy}, c_{sz})\right\}_{s \in \mathcal{S}}$ and post-synapse point labels $P_{\mathcal{S}}=\left\{(p_{sx}, p_{sy}, p_{sz})\right\}_{s \in \mathcal{S}}$, while
only target volumes $X_{\mathcal{T}}=\left\{x_t \in \mathbb{R}^{H \times W \times D}\right\}_{t \in \mathcal{T}}$ are available in the target domain. Note that $H$, $W$, and $D$ denote the
height, width, and depth of volumes, respectively.  The goal is to design a framework that can correctly predict the synapse location for target data $X_\mathcal{T}$.

Fig.~\ref{fig:framework} illustrates the overview of our proposed framework. In our framework, we consider the detection problem as a segmentation task by predicting instance masks for pre- and post-synapses. Specifically, we first generate instance masks for synapses, and then determine the location of the synapse by finding the center of these masks. Since synapse regions are small, we create square masks with a radius of $R$ as ground truth semantic masks for the source data. The pixel-wise one-hot labels for pre-synapses are denoted as $Y_{\mathcal{S}}=\left\{y_s \in\{0,1\}^{H \times W \times D}\right\}_{s \in \mathcal{S}}$, and those for post-synapse are denoted as $M_{\mathcal{S}}=\left\{m_s \in\{0,1\}^{H \times W \times D}\right\}_{s \in \mathcal{S}}$. 

\subsection{Training}
Fig.~\ref{fig:framework} provides an overview of our training framework. The training procedure consists of two stages: (1) training a synapse segmentation network on source data, and (2) generating pseudo labels for target data and fine-tuning the synapse segmentation network on both source and target regions.

\textbf{Segmentation network.}
Our segmentation network, SegNet, outputs two channels for predicting pre-synaptic masks and post-synaptic masks, respectively. As for the network architecture, we follow the superhuman~\cite{lee2017superhuman} to adopt the residual symmetric U-Net to extract the synapse feature. Considering the data we used are imaged using enhanced Focused Ion Beam Scanning Electron Microscope (FIB-SEM) ~\cite{knott2008serial} with an isotropic voxel size, we use isotropic 3D convolutions in the SegNet. Specifically, we ﬁrst embed the feature maps with a $5 \times 5 \times 5$ conventional layer. The modules in other scales always have 3 convolutions with size $3 \times 3 \times 3$. In each module, there is a residual skip connection~\cite{he2016deep} for the second convolution.

\textbf{Pseudo label generation.} The goal of pseudo-labeling is to generate pseudo-labels for unlabeled target data using a model trained on labeled source data~\cite{lee2013pseudo}. We first obtain the segmentation probability outputs for each target volume. The predicted pre-synapse and post-synapse regions are denoted as $\tilde{y}_t$ and $\tilde{m}_t$. With these output probabilities, the initial segmentation pseudo-label for each volume can be generated as
\begin{equation}
\begin{aligned}
\overline{y}_t &=\mathbbm{1}\left[\tilde{y}_t \geq \gamma^{pre}\right] \\
\overline{m}_t &=\mathbbm{1}\left[\tilde{m}_t \geq \gamma^{post}\right] ,
\end{aligned}
\end{equation}
where $\gamma^{pre}, \gamma^{post}$  are the thresholds used to produce hard labels for pre-synapses and post-synapses, respectively. We compute the detection pseudo-labels $\hat{c}_t$ and $\hat{p}_t$ by finding the center location of the initial segmentation pseudo-labels $\overline{y}_t$ and $\overline{m}_t$. Similarly, we generate square masks with radius $R$ based on $\hat{y}_t$ and $\hat{m}_t$, respectively, to obtain the final segmentation labels $\hat{y}_t$ and $\hat{m}_t$ for target data.

\textbf{Loss function.} The binary cross entropy (BCE) is a common loss function used in biomedical image segmentation. The sparsity of synapses leads to severe class imbalance problems between foreground and background for semantic segmentation. Thus, we adopt weighted BCE (WBCE) loss as the segmentation loss. For the first stage, the total loss for SegNet as
\begin{equation}
\begin{aligned}
    \mathcal{L}_{stage1} &=\mathcal{L}^S_{seg}(X_S, Y_S) + \mathcal{L}_{seg}^S(X_S, M_S) \\
    &= \mathcal{L}_{\text{WBCE}} ( \tilde{Y}_S, Y_S) + \mathcal{L}_{\text{WBCE}}( 
    \tilde{M}_S, M_S).
\end{aligned}
\end{equation}
For the second stage, the total loss on for SegNet as
\begin{equation}
\begin{aligned}
    \mathcal{L}_{stage2} =& \mathcal{L}_{\text{WBCE}} ( \tilde{Y}_S, Y_S) + \mathcal{L}_{\text{WBCE}}( 
    \tilde{M}_S, M_S) \\
    + & \mathcal{L}_{\text{WBCE}} ( \tilde{Y}_T, \hat{Y}_T) + \mathcal{L}_{\text{WBCE}}( 
    \tilde{M}_T, \hat{M}_T).
\end{aligned}
\end{equation}

\subsection{Inference}
During inference, we adopt connected component labeling to separate individual synapses in the pre-synapse mask predictions and post-synapse mask predictions. We determine the location of each synapse by calculating the center point coordinates of their respective masks. To match pre-synapses with post-synapses, we assign the nearest pre-synapse ID to each post-synapse with the nearest neighbor principle.

 Due to the large size of the EM volume, we need to infer the EM volume patch by patch using a sliding window approach. However, this method leads to lower accuracy near the borders of each patch. Inspired by ~\cite{lee2017superhuman}, we perform inference on overlapping patches and blend them together using a bump function. In our case, we use 50\% overlap in all three dimensions for better results.
 

\section{EXPERIMENTS}
\subsection{Dataset}
The WASPSYN~\cite{wu2023out} dataset includes 14 image chunks from different brain regions of \textit{Megaphragma viggianii} in three whole-brain datasets. Specifically, there are three chunks from specimen one, three chunks from specimen two, and eight chunks from specimen three. For the source data, five volumes with point annotations are taken from specimen three. The remaining nine chunks are considered target data. The accuracy of predicted synapse locations for target data can be obtained after submitting predictions on the challenge website\footnote{\url{https://codalab.lisn.upsaclay.fr/competitions/9169}}.

\begin{table}[t]
	\centering
	\setlength{\tabcolsep}{2mm} 
	\begin{tabular}{c|ccc}
		\toprule
		Method  &  $\text{F1}_{pre}\uparrow$ &  $\text{F1}_{post}\uparrow$	&  $\textbf{F1}\uparrow$  \\
		\midrule
		Ours &0.7568 &0.4752 & \textbf{0.6160}\\
		2nd & - & - & 0.4917\\
            3rd & -	& - & 0.4077 \\
            4th & -	& - & 0.3086 \\
            5th & -	& - & 0.2749 \\
            6th & -	& - & 0.1291 \\
		\bottomrule
	\end{tabular}
	\caption{WASPSYN Challenge leaderboard for target data.}
	\label{table:leadboard}
\end{table}

\subsection{Implementation Details}
For all experiments, we set the batch size to 1. Additionally, we use the Adam optimizer and implement a step learning rate schedule. The initial learning rate is set to 0.0001, with linear warming up in the first 1000 iterations. We apply a weight decay of 0.05 to all layers. For data augmentation, we follow ~\cite{li2022advanced} and employ the common techniques used for EM images. During training stage 1, we use a crop size of $128 \times 128 \times 128$ and train for 20k iterations. For evaluation, we randomly select a volume from the source data. In stage 2, we fine-tune the best model from stage 1 for an additional 10k iterations. For the hyperparameters in our method, we set $R=3, \gamma^{pre}=0.75,  \gamma^{post}=0.65$  by default. There are 3 volumes from the same brain region as the source data. Thus, we fine-tune the model using both the source data and the target data in stage 2.

\subsection{Evaluation Metrics and Challenge Results}
Following ~\cite{wu2023out}, we use the F1-score as the evaluation metric, which is calculated using true positives (TPs), false positives (FPs), and false negatives (FNs). The detection accuracy is evaluated by solving an assignment problem that minimizes the Euclidean distance between predicted synapses and ground truth synapses to find true matches. The F1-score is then calculated. In stage 2, only when the pre-synapse ID matches the post-synapse ID, it can be considered a true positive (TP).

In the WASPSYN Challenge at ISBI 2023, our method ranks the 1st place. As shown in Table~\ref{table:leadboard}, the proposed method notably outperforms other competitors on target data. 

\begin{table}[t]
	\centering
	\setlength{\tabcolsep}{2mm} 
	\begin{tabular}{c|cccc}
		\toprule
		Task  & $\text{F1}_{pre}\uparrow$ & $\text{F1}_{post}\uparrow$ &  $\text{F1}^{*}_{post}\uparrow$	&  $\textbf{F1}\uparrow$  \\
		\midrule
		detection & 0.8089&0.3205 &0.3935  & 0.5647 \\
		segmentation & \textbf{0.8936} & \textbf{0.3748} & \textbf{0.4739} & \textbf{0.6355}\\
		\bottomrule
	\end{tabular}
	\caption{Comparison results between segmentation task and detection task on the evaluation volume. $F1^{*}_{post}$ represents the F1-score that does not consider the match between pre-synapses and post-synapses.}
	\label{table:seg&detect}
\end{table}

\begin{table}[t]
	\centering
	\setlength{\tabcolsep}{2mm} 
	\begin{tabular}{c|cccc}
		\toprule
		Radius  & $\text{F1}_{pre}\uparrow$ & $\text{F1}_{post}\uparrow$ &  $\text{F1}^{*}_{post}\uparrow$	&  $\textbf{F1}\uparrow$  \\
		\midrule
		1 &0.8442 &0.2937 & 0.3921 &0.5690\\
		2 & 0.8747 & 0.3621 & 0.4649 & 0.6184\\
            3 & \textbf{0.8963}	& \textbf{0.3748} &\textbf{0.4735} &  \textbf{0.6355} \\
            4 & 0.8695	& 0.3203 & 0.3860 & 0.5949 \\
            5 & 0.8566	& 0.3380 & 0.4158 & 0.5973 \\
		\bottomrule
	\end{tabular}
	\caption{Ablation results about radius of square masks on the evaluation volume.}
	\label{table:radius}
\end{table}

\begin{table*}[t]
\fontsize{8.8}{10.8}\selectfont
	\centering
	\setlength{\tabcolsep}{2mm} 
	\begin{tabular}{c|l|cccccccccc}
		\toprule
	Stage& Metric  & Volume 1 & Volume 2 & Volume 3 & Volume 4 & Volume 5 &Volume 6 &Volume 7 &Volume 8 &Volume 9 & \textbf{Mean}  \\
		\midrule
		\multirow{3}{*}{ \uppercase\expandafter{\romannumeral1}}& $\text{F1}_{pre}$ &0.7879 &0.6199 &0.6492 & 0.7120 & 0.6492 & 0.8404 & 0.8421 & 0.8595 & 0.7973 &0.7508 \\
		& $\text{F1}_{post}$ & 0.4045 & 0.3534 & 0.5700 & 0.2539 & 0.1543 & 0.2284 & 0.5495 & 0.6705 & 0.5606 & 0.4162\\
            &F1 & 0.5962& 0.4867 & 0.6096   & 0.4829 & 0.4018 & 0.5344 & 0.6958 & 0.7650 &0.6789 &0.5835 \\ \midrule
        \multirow{3}{*}{\uppercase\expandafter{\romannumeral2}}& $\text{F1}_{pre}$ &0.7961 &0.6503 & 0.6366 & 0.7565 & 0.6242 & 0.8372 & 0.8513 & 0.8507 & 0.8078 & \textbf{0.7568}\\ 
		& $\text{F1}_{post}$ & 0.4505 & 0.4589 & 0.5977 & 0.3961 & 0.2274 & 0.3252 & 0.5472 & 0.6826 & 0.5905 & \textbf{0.4752}\\
            &F1 & 0.6233& 0.5546 & 0.6171 & 0.5763 & 0.4258 & 0.5812 & 0.6993 & 0.7666 & 0.6991 & \textbf{0.6160} \\
		\bottomrule
	\end{tabular}
	\caption{Ablation results of two stage on target data.}
	\label{table:stage}
\end{table*}

\subsection{Ablation Study}
\textbf{Superiority of segmentation.} One intuitive method to detect synapses is through keypoints detection~\cite{yin2021center}. Following the approach described in~\cite{zhou2019objects}, we conduct a baseline using key point detection to predict the location of synapses. The peaks in the heatmap correspond to the centers of synapses. To train the baseline, we use weighted MSE loss to supervise the network to locate these peaks in the heatmap. Table~\ref{table:seg&detect} compares the optimization objectives of segmentation and detection on an evaluation volume for stage 1. The results clearly demonstrate that segmentation outperforms detection significantly. Therefore, we choose to use segmentation as the optimization objective for predicting synapse locations.

\begin{table}[t]
	\centering
	\setlength{\tabcolsep}{2mm} 
	\begin{tabular}{l|cc||l|cc}
		\toprule
		  $\gamma^{pre}$ &  $\text{F1}_{pre}\uparrow$ &  $\text{F1}_{post}\uparrow$&  $\gamma^{post}$&  $\text{F1}_{post}\uparrow$ \\
		\midrule
		0.65 &0.7523 &0.4716 &0.55 & 0.4715\\
		0.7 & 0.7546 &0.4705 & 0.6 & 0.4725\\
            0.75 & \textbf{0.7568}&\textbf{0.4752}	& 0.65 &\textbf{0.4752} \\
            0.8 & 0.7563&0.4742	& 0.7 & 0.4707\\
            0.85 & 0.7514&0.4750	& 0.75 & 0.4652 \\ 
             \bottomrule
	\end{tabular}
	\caption{Ablation results about the threshold for pseudo labeling. When $\gamma_{pre}$ is changed, $\gamma_{post}$ remains fixed at 0.65. Conversely, when $\gamma_{post}$ is changed, $\gamma_{pre}$ remains fixed at 0.75.}
	\label{table:threshold}
\end{table}

\textbf{Radius of square masks.} The accuracy of our ground truth synapse masks is determined by the radius of square masks. Therefore, we conduct an ablation study on the radius of square masks for stage 1. Table~\ref{table:radius} shows that setting $R=3$ achieves the best performance in stage 1.
 
\textbf{Thresholds for pseudo labeling.} The correctness of pseudo labels for target data is determined by the threshold of pseudo labeling. In the finetune stage, we investigate the impact of different thresholds on pseudo-labeling. Table~\ref{table:threshold} demonstrates that when $\gamma_{post}$ is fixed at 0.65 and $\gamma_{pre}$ is varied, the F1-score achieves its highest performance with $\gamma_{pre}=0.75$. Similarly, when $\gamma_{pre}$ is fixed at 0.75 and $\gamma_{post}$ is changed, setting $\gamma_{post}=0.65$ yields the best F1-score. Therefore, we set $\gamma_{pre}=0.75$ and $\gamma_{post}=0.65$ as default values based on these results.

\textbf{Pseudo label generation.}
 Another way to generate pseudo labels is by directly using the segmentation output of target data. We conduct this approach as a baseline and set $\gamma_{pre}=0.75$ and $\gamma_{post}=0.65$ to binary segmentation output for generating pseudo labels. The comparison results in Table~\ref{table:pseudo label way} show that our method using square masks is superior.

 \begin{table}[t]
	\centering
	\setlength{\tabcolsep}{2mm} 
	\begin{tabular}{l|ccc}
		\toprule
		 Pseudo label generation &  $\text{F1}_{pre}\uparrow$ &  $\text{F1}_{post}\uparrow$& $\textbf{F1}\uparrow$ \\
		\midrule
		Segmentation outputs &0.7469 &\textbf{0.4763} &0.6116
  \\
  Square masks &\textbf{0.7568} &0.4752 &\textbf{0.6160} \\
             \bottomrule
	\end{tabular}
	\caption{Ablation results about the way of pseudo label generation.}
	\label{table:pseudo label way}
\end{table}

\begin{figure}[t]
    \centering
    \includegraphics[width=\linewidth]{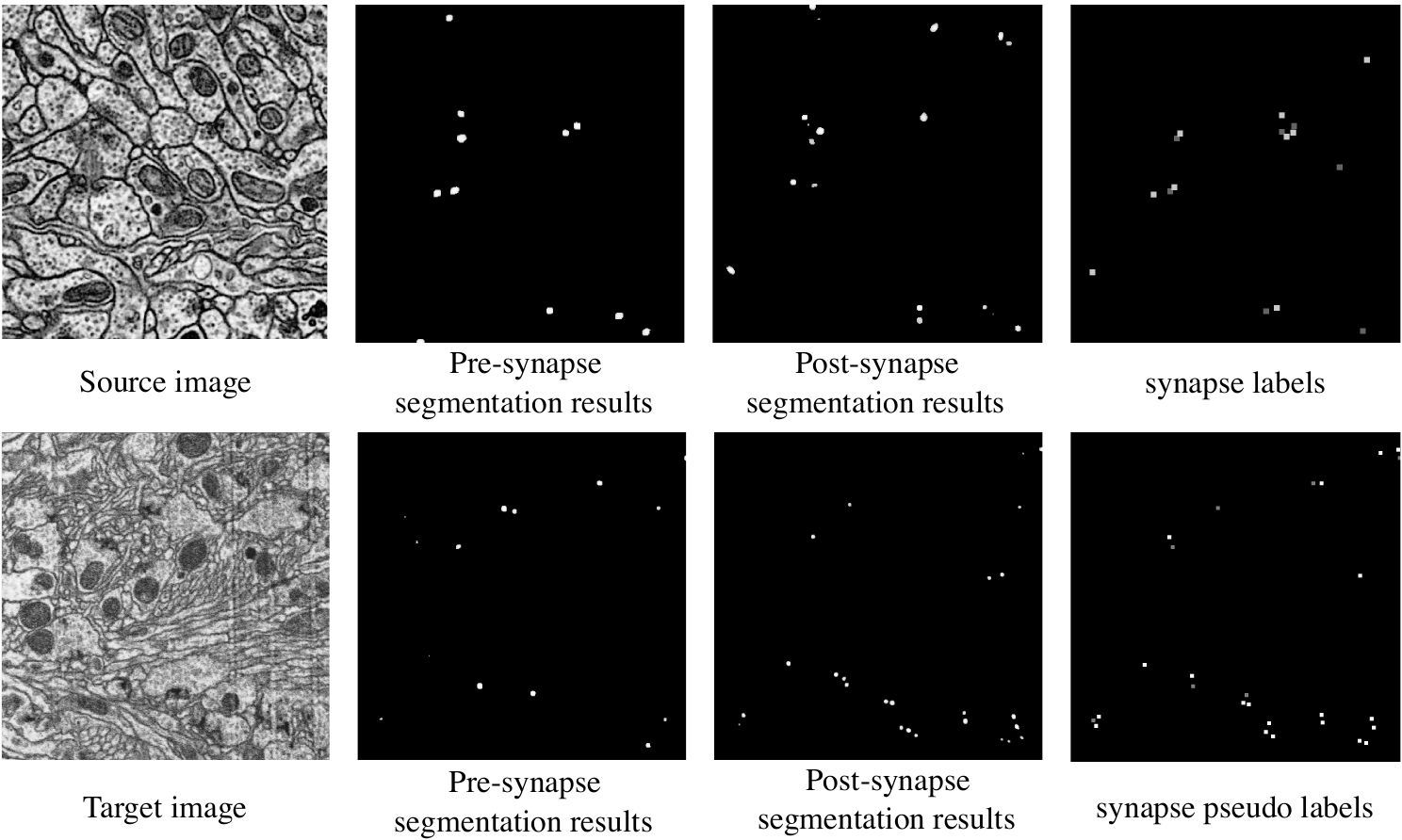} 
    \vspace{-6mm}
    \caption{Visualized results on source data and target data. The white masks represent pre-synapse regions, while the gray masks indicate post-synapse regions.}
    \label{fig:visualization}
\end{figure}

\textbf{Two stages.}
We conduct an ablation study for the two stages to validate their effectiveness. The results presented in Table~\ref{table:stage} demonstrate that the model in stage 1 performs well in generalizing to target data for pre-synapse segmentation. Furthermore, fine-tuning on target data using pseudo labels significantly improves the F1-score of post-synapses.

\textbf{Visualization.} We present visualization results of the source data and target in Fig~\ref{fig:visualization}. It is evident that our proposed method accurately identifies the majority of synapses.

\section{Conclusion}
In this paper, we present AdaSyn, a two-stage segmentation-based framework for domain adaptive synapse detection with weak point annotations. Speciﬁcally, we address the detection problem by utilizing a segmentation-based pipeline to obtain synaptic instance masks. Additionally, we improve model generalizability on target data by regenerating square masks to get high-quality pseudo labels. Benefiting from our high-accuracy detection results, we use the distance nearest principle to match paired pre-synapses and post-synapses. In the WASPSYN challenge at ISBI 2023, our method ranks the 1st place.


\bibliographystyle{IEEEbib}
\bibliography{refs}

\end{document}